# Unveiling the mystery of visual information processing in human brain

Emanuel Diamant

VIDIA-mant, P.O. Box 933, 55100 Kiriat Ono, Israel

emanl@012.net.il

**Abstract.** It is generally accepted that human vision is an extremely powerful information processing system that facilitates our interaction with the surrounding world. However, despite extended and extensive research efforts, which encompass many exploration fields, the underlying fundamentals and operational principles of visual information processing in human brain remain unknown. We still are unable to figure out where and how along the path from eyes to the cortex the sensory input perceived by the retina is converted into a meaningful object representation, which can be consciously manipulated by the brain. Studying the vast literature considering the various aspects of brain information processing, I was surprised to learn that the respected scholarly discussion is totally indifferent to the basic keynote question: "What is information?" in general or "What is visual information?" in particular. In the old days, it was assumed that any scientific research approach has first to define its basic departure points. Why was it overlooked in brain information processing research remains a conundrum. In this paper, I am trying to find a remedy for this bizarre situation. I propose an uncommon definition of "information", which can be derived from Kolmogorov's Complexity Theory and Chaitin's notion of Algorithmic Information. Embracing this new definition leads to an inevitable revision of traditional dogmas that shape the state of the art of brain information processing research. I hope this revision would better serve the challenging goal of human visual information processing modeling.

## 1 Introduction

Human vision is generally considered as an extremely powerful information processing system that facilitates our successful interaction with the surrounding world. Near a half of our cerebral cortex is busy with processing visual information [1]. However, and despite of extended and extensive research efforts, which encompass many different fields of exploration, the underlying fundamentals and operational principles of human visual information processing remain a puzzle and a conundrum for many generations of thinkers, philosophers, and scientific researchers.

Nevertheless, a working theory of human visual information processing has been established few decades ago by the seminal works of David Marr [2], Anne Treisman [3], Irving Biederman [4], and a large group of their associates and followers. The theory considers human visual information processing as an interplay of two inversely directed processing streams. One is an unsupervised, bottom-up directed process of initial image information pieces discovery and localization. The other is a supervised, top-down directed process, which conveys the rules and the knowledge that guide the linking and binding of these disjoint information pieces into perceptually meaningful image objects.

Essentially, as an idea, this proposition was not entirely new. About two hundred years ago, Kant had depicted the "faculty of (visual) apperception" as a "synthesis" of two constituents: the raw sensory data and the cognitive "faculty of reason" [5]. A century later, Herman Ludwig Ferdinand von Helmholtz (the first who scientifically investigated our senses) had reinforced this view, positing that sensory input and perceptual inferences are different, yet inseparable, faculties of human vision [6]. The novelty of the modern approach was in an introduction of a new concept used for the idea clarification - "visual information" [2]. However, a suitable definition of the term was not provided, and the mainstream of relevant biological research has continued (and continues today) to investigate the puzzling duality of the phenomenon by capitalizing on traditional vague definitions of the matters: local and global image content, perceptual and cognitive image processing, low-level computer-derived image features versus high-level human-derived image semantics [7], [8]. Putting aside the terminology, the main problem of human visual information processing remains the same: to fulfill the intuitively effortless low-level information pieces agglomeration into meaningful semantic objects the system has to be provided with some high-level knowledge about the rules of this agglomeration. Needless to say, such rules are usually not available. In biological vision research, this predicament is known as the "binding problem". Its importance was recognized at very early stages of vision research, and massive efforts have been directed into it in order to reach a suitable and an acceptable solution. Despite the continuous efforts, unfortunately, any discernable success was not achieved yet. (For more details, see [9] and a special issue of Neuron, vol. 24, 1999, entirely devoted to the problem).

Unable to reach the required high-level processing (binding) rules, vision research took steps in a previously forbidden, but now supposedly the only one possible direction – to try to derive the needed high-level knowledge from the available low-level information pieces. A rank of theoretical and experimental work has been done in order to support and justify this just-mentioned shift in research aspirations. Two approaches could be distinguished in this regard: chaotic attractor modeling approach [10], [11], and saliency attention map modeling approach [12], [13]. I will not review the details of these approaches here. I will only make a note that both of them presume low-level bottom-up processing as the most proper way for high-level information recovery [25], [26]. Both are computationally expensive. Both definitely violate the basic assumption about the leading role of high-level knowledge in the low-level information processing.

Such violation can occur only in an atmosphere of total indifference to the theory's fundamental claims about high-level information superiority in the general course of visual information processing. Subsequently, such indifference stems from a very loose understanding about what is the concept of "information", what is the right way to use it properly, and what information treatment options could arise from this understanding.

## 2 Attempting to define "What is information?"

We live today in an Information Age, where information is an indispensable ingredient of our life. We consume it, create it, seek for it, transfer, exchange, hide, reveal, accumulate, and disseminate it – in one word: information is a remarkably important part of our life. But can someone explain to me what we have in mind when the term "information" is used? My attempts (undertaken several years ago) to get my own answer for this question were so desperate that I was almost ready to accept the stance that information is an indefinable entity, like, for example, "space" and "time" in classical physics. (A similar confession I have found in a recent paper of A. Sloman [14], where he compares the indefinable notion of "information" with the indefinable notion of "energy".)

Fortunately, I have eventually hit on an information definition fitting my visual information processing needs. It turns out that this definition can be derived from Solomonoff's theory of Inference [15], Chaitin's Algorithmic Information theory [16], and Kolmogorov's Complexity theory [17]. Recently, I have learned that Kolmogorov's Complexity and Chaitin's Algorithmic Information theory are referred as respected items of a list of seven possible contestants suitable to define what is information [18]. In this regard, I was very proud of myself that I was lucky to avoid the traps of Shannon's Information Theory, which is known to be useful in communication applications, but it is absolutely inappropriate for visual information explorations that I am trying to conduct. The reason for this is that Shannon's information properly describes the integrated properties of an information message, while Kolmogorov's definition is suitable for evaluation of information content of separate isolated message subparts (separate message objects). This is, indeed, much closer to the way in which humans perceive and grasp their visual information.

The results of my investigation have been already published on several occasions, [19], [20], [21], [22], and interested readers can easily get them from a number of freely accessible repositories (e.g., arXiv, CiteSeer (the former Research Index), Eprintweb, etc.). Therefore, I will only repeat here some important points of these early publications, which properly reflect my current understanding of the matters.

The main point is that **information is a description**, a certain alphabet-based or language-based description, which Kolmogorov's theory regards as a program that (being executed) trustworthy reproduces the original object [23]. In an image, such objects are visible data structures from which an image consists of. So, a set of reproducible descriptions of image data structures is the information contained in an image.

The Kolmogorov's theory prescribes the way in which such descriptions must be created: at first, the most simplified and generalized structure must be described. Then, as the level of generalization is gradually decreased, more and more fine-grained image details (structures) are become revealed and depicted. This is the second important point, which follows from the theory's pure mathematical considerations: image **information is a hierarchy of recursive decreasing level descriptions** of information details, which unfolds in a coarse-to-fine top-down manner. (Attention, please: any bottom-up processing is not mentioned here. There is no low-level feature gathering and no feature binding. The only proper way for image information elicitation is a top-down coarse-to-fine way of image processing.)

The third prominent point, which immediately pops-up from the two just mentioned above, is that the top-down manner of image **information elicitation does not require incorporation of any high-level**

**knowledge** for its successful accomplishment. It is totally free from any high-level guiding rules and inspirations. That is why we have named this information **The Physical Image Information**, which is independent of any high level interpretation of it.

What immediately follows from this is that high-level image semantics is not an integrated part of image information content (as it is traditionally assumed). It cannot be seen more as a natural property of an image. Image semantics must be seen as a property of a human observer that watches and scrutinizes an image. That is why we can say today: **semantics is assigned to an image by a human observer**. That is strongly at variance with the contemporary views on the concept of semantic information. Following the new information elicitation rules, it is impossible to continue to pretend that semantics can be **extracted from an image**, (as in [24]), or should be **derived from low-level information features** (as in [25], [26] and many other related publications). That simply does not hold any more.

## 3   The old problems in a new light

My personal research interests and life-long experience come from the field of computer vision R&D. Computer vision is a sub-field of Artificial Intelligence, which puts creation of intelligent machines with human-like visual performances as its primary goal. Because human vision outperforms any known computer vision implementation, it is only natural for computer vision designers to eagerly follow the progress in human vision research. The inability of human vision research to provide computer vision designers with the necessary leads has forced the latter to find their own ways of problem solving. But at the end, the majority of computer vision designs simply replicate the slips and the misconceptions of human vision research. Our case is a remarkable exception from this custom. However, to verify the validity of our findings we would like to return to the well-known and friendly native fields of computer vision design.

Since the new definition of information has forced us to reconsider the traditional philosophy of image information processing, we have chosen the framework of visual robot design as a test-bed for verification of our new ideas. In this enterprise, we are aimed on creating an artificial vision system with some human-like cognitive capabilities. The proposed block-scheme of our solution of the problem is depicted in Fig. 1.

As follows from the preceding discussion, the block-scheme is comprised of two independent but tightly coupled parts: Physical Information processing part and Semantic Information processing part.

It is generally agreed that the first stage of any system has to be an image segmentation stage at which the whole bulk of image pixels (image raw data) has to be decomposed into a finite set of image patches (in terms of our approach – the Physical Information content of an image has to be extracted). Afterwards, the segmented pieces are submitted to a process of image analysis and interpretation (in terms of our approach – Semantic Information has to be assigned to the image).

As one can see, the proposed Physical Information extraction part is comprised of three main processing units: the bottom-up processing path, the top-down processing path and a stack where the discovered information content (the generated descriptions of it) are actually accumulated. (More details about Physical Information extraction can be found in [19], [20], [21], [22]).

As follows from the information extraction principles (which prescribe that the most general and simplified descriptions have to be derived first), the purpose of the bottom-up processing path is to provide a simplified compressed copy of an input image. The original image is squeezed along this path to a small size of approximately 100 pixels. The rules of the shrinking operation are very simple and fast: four non-overlapping neighbor pixels in an image at level $L$ are averaged and the result is assigned to a pixel in a higher ($L+1$)-level image, (a so-called 4 to 1 image compression). At the top of the shrinking pyramid, the image is segmented, and each segmented region is labeled. Since the image size at the top is significantly reduced and since in course of the bottom-up image squeezing a severe data averaging is attained, the image segmentation/classification procedure does not demand special computational resources.

From this point on, the top-down processing path is commenced. At each level, the segmentation maps are expanded to the size of an image at the nearest lower level, (a 1 to 4 expansion). Since the regions at different hierarchical levels do not exhibit significant changes in their characteristic intensity, the majority of newly assigned pixels are determined in a sufficiently correct manner. Only pixels at region borders and seeds of newly emerging regions may significantly deviate from the assigned values. Taking the corresponding current-level image as a reference (the left-side unsegmented image), these pixels can be easily detected and subjected to a refinement cycle. In such a manner, the process is subsequently repeated at all descending levels until the segmentation of the original input image is successfully accomplished.

At each processing level, every image object-region (whether just recovered or an inherited one) is registered in the objects' appearance list (the Stocked Level Descriptions rectangle in Fig. 1), which is the third constituting part of the proposed scheme.

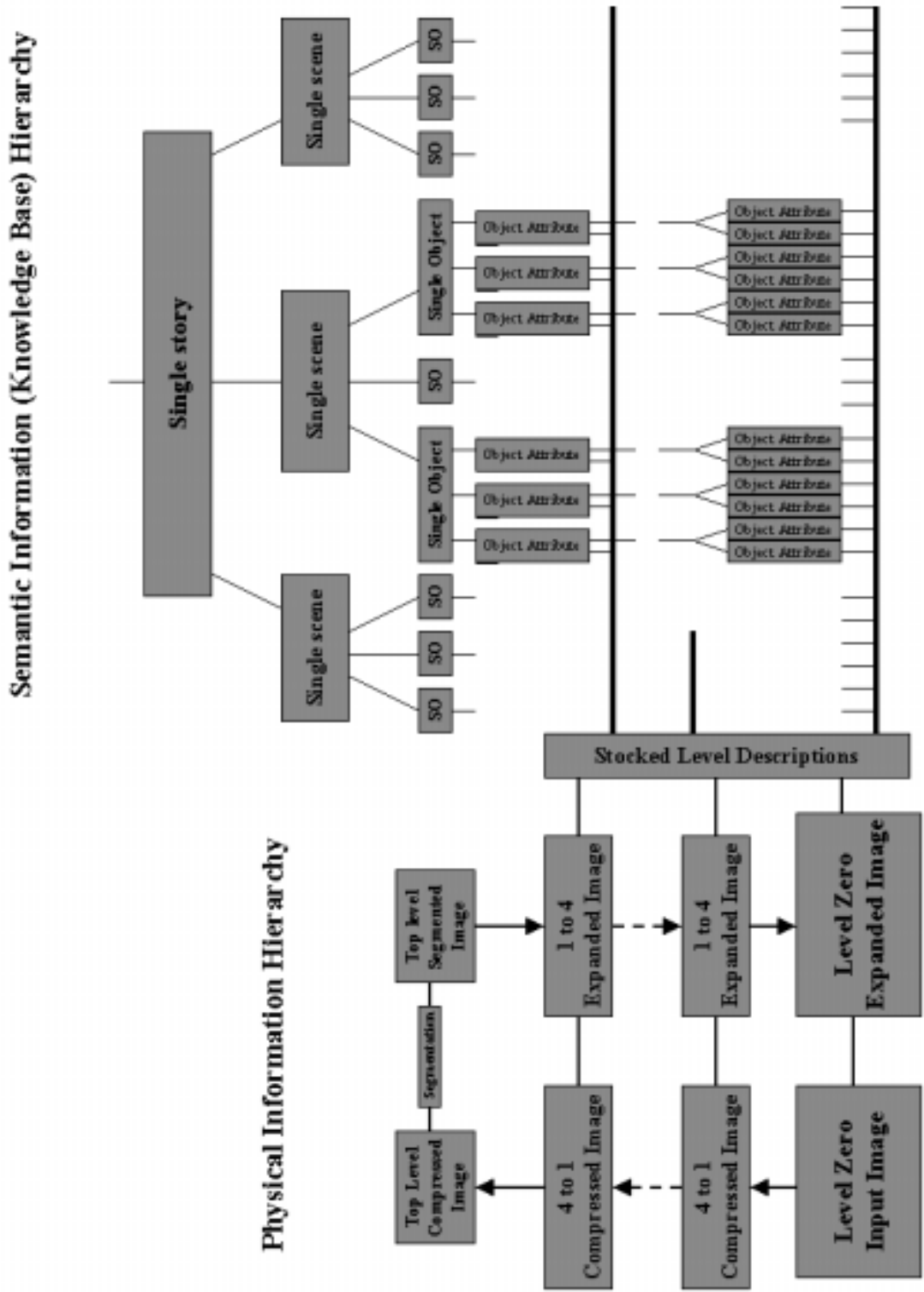

Fig. 1. Arrangement of Physical and Semantic Information Hierarchies and their interconnection.

The registered object parameters are the available simplified object's attributes, such as size, center-of-mass position, average object intensity and hierarchical and topological relationship within and between the objects ("sub-part of…", "at the left of…", etc.). They are sparse, general, and yet specific enough to capture the object's characteristic features in a variety of descriptive forms.

This way, a practical algorithm based on the announced above principles has been developed and subjected to some systematic evaluations. The results were published, and can be found in [19], [20], [21], [22]. There is no need to repeat again and again that excellent, previously unattainable segmentation results have been attained in these tests, undoubtedly corroborating the new information processing principles. Not only an unsupervised segmentation of image content has been achieved, (in a top-down coarse-to-fine processing manner, without any involvement of high-level knowledge), a hierarchy of descriptions for each and every segmented lot (segmented object) has been achieved as well. It contains a set object related parameters (object related **information**), which enable subsequent object reconstruction. That is exactly what we have previously defined as information. That is the reason why we specify this information as "physical information", because that is the only information present in an image, and therefore **the only information that can be extracted from an image**. For that reason it is dissociated from the semantic information, which (as we understand now) express a property of an external observer. Therefore it must be treated (or modeled) in accordance with specific his/her cognitive information processing rules.

What are these rules? A consensus view on this topic does not exist as yet in the biological vision theories as well as in the computer vision practice. So, we have to blaze our own trails. We decided, thus, to meet this challenge by suggesting a new approach based on our previously declared information elicitation principles. The preliminary results of our first attempt were published recently in [27]. As in the case of physical information, we will not repeat here all the details of this publication. A possible solution for the Semantic Information processing part is depicted in Fig. 1. Here we will proceed only with a brief explanation of some of its critical points.

Human's cognitive abilities (including the aptness for image interpretation and the capacity to assign semantics to an image) are empowered by the existence of a huge knowledge base about the things in the surrounding world kept in human brain/head. This knowledge base is permanently upgraded and updated during the human's life span. So, if we intend to endow our visual robot with some cognitive capabilities we have to provide it with something equivalent to this (human) knowledge base.

It goes without saying that this knowledge base will never be as large and developed as its human prototype. But we are not sure that such requirements are valid in our case. After all, humans are also not equal in their cognitive capacities, and the content of their knowledge bases is very diversified too. (The knowledge base of aerial photographs interpreter is certainly different from the knowledge base of X-ray images interpreter, or IVUS images, or PET images). The knowledge base of our visual robot has to be small enough to be effective and manageable, but sufficiently large to ensure the robot's acceptable performance. Certainly, for our feasibility study we can be satisfied even with a relatively small, specific-task-oriented knowledge base.

The next crucial point is the knowledge (base) representation issue. To deal with it, we first of all must arrive at a common agreement about what is the meaning of the term "knowledge". (A question that usually has no commonly accepted answer.) We state that in our case a suitable and a sufficient definition of it would be: "Knowledge is a memorized information". Consequently, we can say that knowledge (like information) must be a hierarchy of descriptive items, with the grade of description details growing in a top-down manner at the descending levels of the hierarchy.

What else must be mentioned here, is that these descriptions have to be implemented in some alphabet (as it is in the case of physical information) or in a description language (which better fits the semantic information case). Any farther argument being put aside, we will declare that the most suitable language in our case is a natural human language. After all, the real knowledge bases that we are familiar with are implemented in a natural human language.

The next step, then, is predetermined: if natural language is a suitable description implement, the suitable form of this implementation is a narrative, a story tale [28]. If the description hierarchy can be seen as an inverted tree, then the branches of this tree are the stories that encapsulate human's experience with the surrounding world. And the leaves of these branches are single words (single objects) from which the story parts (single scenes) are composed of.

The descent into description details, however, does not stop here, and each single word (single object) can be farther decomposed into its attributes and rules that describe the relations between the attributes.

At this stage the physical information enters the game. Because the words are usually associated with physical objects in the real world, words' attributes must be seen as memorized physical information (descriptions). Once derived (by a visual system) from the observable world and learned to be associated with a particular word, these physical information descriptions are soldered in into the knowledge base. Object recognition, thus, turns out to be a comparison and similarity test between currently acquired physical information and the one already retained in the memory. If the similarity test is successful, starting from this point in the hierarchy and climbing back up on the knowledge base ladder we will obtain: first, the linguistic label for a recognized object; second, the position of this label (word) in the context of the whole story; and third, the ability to verify the validity of an initial guess by testing the appropriateness of the neighboring parts composing the object or the context of a story. In this way, object's meaningful categorization can be reached, and the first stage of image annotation can be successfully accomplished, providing the basis for farther meaningful (semantic) image interpretation.

One question has remained untouched in our discourse: How this artificial knowledge base has to be initially created and brought into the robot's disposal? The vigilant reader certainly remembers the fierce debates about learning capabilities of neural networks and other machine learning technologies. We are aware of these debates. But in our case we can state certainly: they are irrelevant. For a simple reason: the top-down fashion of the knowledge base development pre-determines that all responsibilities for knowledge base creation have to be placed on the shoulders of the robot designer.

Such an unexpected twist in design philosophy will be less surprising if we recall that human cognitive memory is also often defined as a "declarative memory". And the prime mode of human learning is the declarative learning mode, when the new knowledge is explicitly transferred to a developing human from his external surrounding: From a father to a child, from a teacher to a student, from an instructor to a trainee. So, our proposal that robot's knowledge base has to be designed and created by the robot supervisor is sufficiently correct and is fitting our general concept of information use and management.

## 4  Human vision related issues

The introduction of a new information definition has its important consequences also for brain information processing understanding. First of all, the classical theory of interacting bottom-up and top-down processing streams cannot be regarded any more as trust-worthy and correct. Consequently, all new theories that are built on its basic assumptions have serious problems explaining and justifying their experimental evidence. From the point of view of these theories, Physical Information processing (they associate it usually with feed-forward bottom-up processing) and Semantic Information processing (they associate it with top-down task-dependent mediation) are still tightly entangled and intertwined streams of brain visual information processing. As a result, the most of the research work is busy with attempts to associate the imaginary processing pathways with anatomic structures discernible in the brain [29], [30]. (I provide here these two references only for illustration purposes. In the real world the number of this sort of publications is uncountable).

While the mainstream of human vision research continues to approach visual information processing in a bottom-up fashion, it turns out that the idea of primary top-down processing was never extraneous to biological vision. The first publications addressing this issue are dated by the early eighties of the last century, (David Navon at 1977 [31], and Lin Chen at 1982 [32]). The prominent authors were persistent in their claims, and farther research reports were published regularly until the recent time, [33], [34]. However, it looks like they have been overlooked, both in biological and in computer vision research. Only in the last years, a tide of new evidence has become visible and is pervasively discussed now. Although the spirit of these discussions is still different from our view on the subject, the trend is certainly in favor of the foremost top-down visual information processing [35], [36]. (Again, top-down information processing in the physical information processing part only is assumed here. Information processing partition proposed in this paper is not acknowledged by the contemporary vision researchers.)

Defining information as a description message has another far-reaching consequence. If this assumption is right, then the current belief that a spiking neuron burst is a valid form of information exchange and representation [37] does not hold any more. The variance in spikes' heights or duty times is an inadequate alphabet to implement information descriptions of a desired complexity. We can boldly speculate that a biomolecular alphabet would be a much better and appropriate solution in such a case. Support for this kind of speculations can be derived from the recent advances in molecular biology research [38], [39] and from the flood of papers about molecular communication peculiarities [40].

For those to whom the idea of molecular code (molecular alphabet) used to compose information descriptions predestined for brain cell communications seems very extravagant and exaggerated, I would like to propose some extended quotations fetched from the recently emerged publications. The subject of these publications is astrocytes, the dominant glial cells, which "listen and talk" [41] with neuronal and synaptic networks. Here is what [42] writes on this occasion: "One reason that the active properties of astrocytes have remained in the dark for so long relates to the differences between the excitation mechanisms of these cells and those of neurons. Until recently, the electrical language of neurons was thought to be the only form of excitation in the brain. Astrocytes do not generate action potentials, they were considered to be non-excitable and, therefore, unable to communicate. The finding that astrocytes can be excited non-electrically has expanded our knowledge of the complexity of brain communication to an integrated network of both synaptic and non-synaptic routs". [43] logically continues this conjecture: "Synaptic signalling between neurons is the basis of information transfer, and the modification of synaptic processes underlies learning and memory. Astrocytes possess many of the same molecules involved in the neuronal synaptic machinery and can influence both synaptic strengthening and depression". And [44] finally specifies the astrocyte-neuron communication means: "Activated astrocytes have the ability to release a variety of neuroactive molecules including glutamate, ATP, nitric oxide, prostaglandins, atrial natriuretic peptide (ANP), and D-serine, which in turn influences neuronal excitability. This bi-directional signalling between astrocytes and neurons has led us to propose that the astrocyte represents a third active element of the synapse together with the pre- and postsynaptic terminals in what we have termed the "tripartite synapse".

Interested readers could continue by themselves to scrutinize the flood of the papers related to the subject.

On the other hand, seeing information descriptions as molecularly encoded messages fits very well other aspects of brain information processing, which are unknown or very superficially known today to the brain researchers. By information processing we mean information storage and retrieval (which presumes information memorization), information transfer and exchange (the case was discussed just above), information transformation (including message concatenation, contraction, as well as insertion and deletion of message parts, and so on). All these information processing tasks presume existence of an underlying effective memory system. The nearest known to us prototype of such a system is the computer memory construction. Thus, if information is an encoded description, memory-based information processing issues could be easily reduced to a trivial task of memory handling, which closely resembles computer memory organization and management. And we can boldly speculate in this regard that dendrite spines are a proper biological wet-ware suitable for such a memory implementation.

By the way, it turns out that computer memory handling terminology is not at all extraneous to brain vision research. Hypotheses about "object files" [45] and "event files" [46] are repeatedly emerging in human vision literature during the last decades. At this point, we would claim that the analogy with computer memory could be even farther extended: the memory structure supporting the semantic knowledge base (which is tightly interconnected with the physical information processing structure) can be seen as computer central processor's cache memory that keeps only the currently used memory files, while the rest of the files are stored in a general memory repository. In terms of biological vision, the cache memory can be seen as the Short-term (Working) memory reproduction, and, respectively, the main computer memory as the Long-term memory prototype. From time to time, the contents of the Short-term memory is being replaced by the relevant files from the Long-term memory, in accordance with the executed tasks needs.

It would be interesting to notice that event file concept also fits very well the narrative knowledge transfer and representation hypothesis proposed earlier in Section 3. In this light, our claim about story-based Semantic Information Knowledge Base creation seems to be as never before plausible. We still don't know who is that ancient writer, the genetic Lev Tolstoy, who diligently compiles his "War and Peace" saga inside a given nervous cell. What we do know certainly is that composing sub-stories of this narrative (including whole sentences and single words) are brought from the outside, from the external world, prepared and ready to be integrated into the new-born masterwork. We have early delineated this process in Section 3, where we have claimed that top-down fashion of Semantic Knowledge Base creation presupposes that knowledge base building information messages have to be brought from the outside. Here we would like to disclose some supporting examples that the nature has endowed us with: Ants that learn in tandem, transferring between them the necessary knowledge [47] (certainly, chemically encoded); bees that convey to the rest of the hive the information about melliferous sites [48]. Even bacteria exhibit a knowledge sharing ability, when a single DNA fragment of one bacteria is disseminated among other colony members facilitating thus their resistance to antibiotic treatment [49]. Exchange of chemically encoded information messages (pheromones) is, obviously, the most known and universal feature common to animals and many other living creatures.

Another brain information processing attribute – oscillatory changes in scalp electro-encephalograms and magneto-encephalograms – can be better explained and understood relying on the information processing principles that we have introduced just above and a possible implementation of which is depicted in a block-scheme in Fig. 1. As follows from our previous explanations (in Section 3), image-understanding process, which encompasses scene recognition sub-tasks, is reified as a sequence of recursive iterative loops in which the consistency of attribute-object-scene subparts is verified before the right set of contributing components is finally determined. We can speculate that this loop-based search is responsible for the observed electro-encephalography and magneto-encephalography oscillations.

This is a brief account of brain information-processing issues that could benefit from our new definition of what information is. If our definition of information as a description is correct, then even more exciting findings concerning visual information processing in human brain could be expected in the near future. I am definitely excited by this opportunity.

## 5  Some concluding remarks

In this paper, we have proposed a new definition of information that is first of all suitable for our computer vision design purposes. We have offered an exploration of benefits that a skilled use of this definition can provide for the human vision research.

The present proposal is incomplete and tentative since this is just a first step, and further research remains to be done. We are aware that our approach is very different from those that are extensively explored and developed in frame of other research programs [50]. However, the enterprise that we are aimed at is not a task for a single person or a small group of developers. It requires consolidated efforts of many interested parties. We hope that the time for this is not far away.